# Physics-informed Machine Learning for Battery Pack Thermal Management

Zheng Liu[#], University of Illinois Urbana-Champaign

Yuan Jiang[#], University of Illinois Urbana-Champaign

Yumeng Li, PhD, University of Illinois Urbana-Champaign

Pingfeng Wang, PhD, University of Illinois Urbana-Champaign

[#] Zheng Liu and Yuan Jiang are co-first authors.

## ABSTRACT

With the popularity of electric vehicles, the demand for lithium-ion batteries is increasing. Temperature significantly influences batteries' performance and safety. Battery thermal management systems can effectively control the temperature of batteries; therefore, batteries' performance and safety can be ensured. However, the development process of battery thermal management systems is time-consuming and costly due to the extensive training dataset needed by data-driven models requiring enormous computational costs for finite element analysis. Therefore, a new approach to constructing surrogate models is needed in the era of AI. Physics-informed machine learning enforces the physical laws in surrogate models, making it the perfect candidate for estimating battery pack temperature distribution. In this study, we first developed a 21700 battery pack indirect liquid cooling system with cold plates on the top and bottom with thermal paste surrounding the battery cells. Then, the simplified finite element model was built based on experiment results. Due to the high coolant flow rate, the cold plates can be considered as constant temperature boundaries, while battery cells are the heat sources. The physics-informed convolutional neural network served as a surrogate model to estimate the temperature distribution of the battery pack. The loss function was constructed considering the heat conduction equation based on the finite difference method. The physics-informed loss function helped the convergence of the training process with less data. As a result, the physics-informed convolutional neural network showed more than 15% improvement in accuracy compared to the data-driven method with the same training data.

## 1 INTRODUCTION

The demand for lithium-ion batteries is increasing with the rising popularity of transportation electrification and the advances in battery materials [1]. However, temperature plays a crucial role in influencing the performance and safety of batteries. Consequently, thermal management systems are essential to maintain optimal temperature to ensure the safety and efficiency of batteries [2], [3]. Thus, battery thermal management systems are needed to enhance safety and performance by effectively controlling battery temperature [4]. Despite their benefits, the development of battery thermal management systems is often time-consuming and expensive [5]. Researchers have utilized finite element models and data-driven methods to accelerate the design process to create surrogate models for estimating battery pack temperature distribution [6]. Nevertheless, data-driven models require extensive training datasets, which incur significant computational costs for finite element analysis [7]. To develop battery thermal management systems effectively, a new approach is needed with the help of the recent breakthrough in artificial intelligence. Physics-informed machine learning, which incorporates physical laws into surrogate models, emerges as an ideal solution for accurately estimating the temperature distribution of the battery [8].

This study adopted a battery pack with indirect liquid cooling, which has been widely applied in industrial applications to cool the battery pack while reducing the manufacturing cost-effectively. Based on previous experiment testing, a simplified finite element model was built to predict the battery pack's temperature accurately. To further reduce the computational cost of the temperature prediction while maintaining accuracy, a physics-informed convolutional neural network was applied to serve as a surrogate model to estimate the temperature distribution of the battery. Different from the data-driven method, the loss function in the physics-informed convolutional neural network was constructed considering the heat transfer equations based on the finite difference method. The physics-informed convolutional neural network achieved higher accuracy than the data-driven method with the same training data. Using the surrogate model based on a physics-informed convolutional neural network, the temperature distribution of the battery can be predicted.

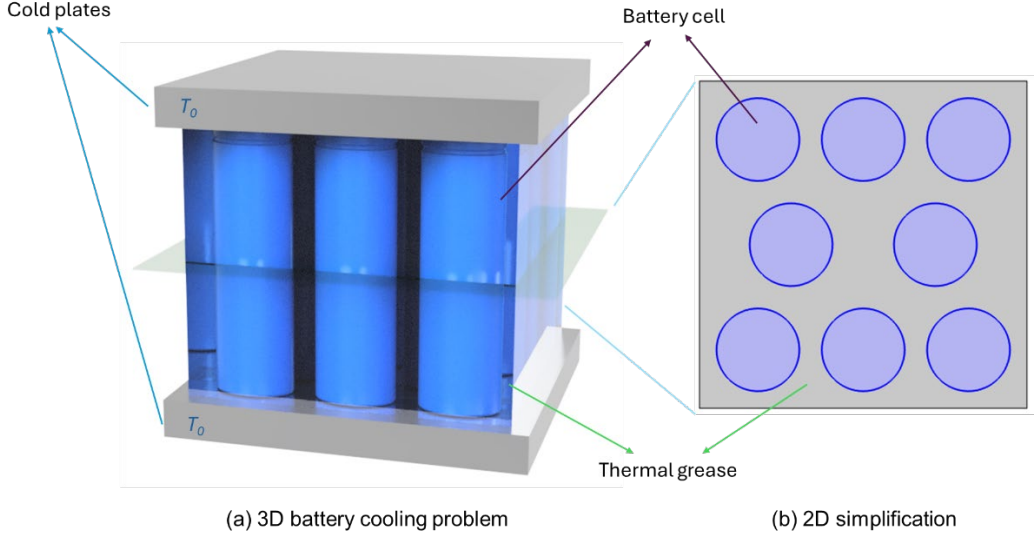

*Figure 1. Battery pack indirect liquid cooling with cold plates simplification.*

## 2 PROBLEM STATEMENT

### 2.1 Finite Element Model Simplification

The geometry of the battery pack must be carefully designed to control its temperature while ensuring its performance effectively. Since the traditional experimental method is time-consuming and expensive, the new method is needed to speed up evaluating new designs. Based on our previous experiment testing, a high-fidelity FE model has been built in COMSOL Multiphysics [9]. Based on the uniformity of the temperature distribution on the battery surface, a simplified 2D axisymmetric model was adopted. This high-fidelity FE model can perfectly represent the heat generation during the battery (INR21700-50E, Samsung SDI) discharging progress as a time-dependent problem. However, the simulation process is tedious since the high-fidelity FE model considers the chemical reaction with the change of time. Moreover, the different material layers contribute to the simulation's complexity. To further accelerate the simulation process for the broader application of batteries, additional simplification for the FE model is needed.

To further simplify the problem, we combined the different layers inside the battery together to reduce the computational cost. The simplified geometry can greatly reduce the computational cost, and it has been validated in our previous work [9], [10], [11]. After analyzing the FE model results, we found that the heat generation rate is relatively stable during the discharging process. Thus, we converted the time-dependent problem to a stationary problem with a constant heat generation rate.

For the indirect liquid cooling for the cylindrical cell battery pack, one effective way is to put the cold plates on the top and bottom of different layers of batteries, as shown in Fig. 1(a). The coolant passes through the channels inside the cold plates, and while the coolant flow rate is sufficient, the temperature of the cold plates will remain the same during the battery pack discharging process. Thermal grease (Parker Chomerics T670, Parker-Hannifin Corporation) is applied to enhance heat transfer between battery cells. Since the cold plates are on the top and bottom of the battery pack, the middle part of the battery will have a higher temperature due to the relatively long distance to the cold plates. Thus, the 2D model focuses on the middle part of the battery cells where the high temperature is more likely.

### 2.2 Finite Element Model Setup

The middle layer of the battery cells within the battery thermal management system using cold plates and thermal grease is illustrated in Fig. 1(b). The battery cells are the heat source, with a volumetric heat generation rate of 176405 W/m$^3$. While thermal grease can absorb heat and transfer to the cold plates, it has a volumetric heat generation rate of $-42857.14(T - T_0)$ W/m$^3$. In which $T_0$ is the temperature of the cold plates. The thermal insulation boundary condition is used, while the initial temperature and the cold plates' temperature are set as 25 °C.

The density and heat capacity of the battery and thermal grease are considered in the simulation. The heat transfer in solid as a stationary problem can be described as

$$\rho C_p \left(\frac{\delta T}{\delta t} + \boldsymbol{u} \cdot \nabla T\right) + \nabla \cdot (\boldsymbol{q} + \boldsymbol{q}_r) = -\alpha T : \frac{dS}{dt} + Q \quad (1)$$

where $\rho$ is the density, $C_p$ is the specific heat capacity at constant stress, $T$ is the absolute temperature, $\boldsymbol{u}$ is the velocity vector of translational motion, $\boldsymbol{q}$ is the heat flux by conduction, and $\boldsymbol{q}_r$ is the heat flux by radiation. $\alpha$ is the coefficient of thermal expansion, $\boldsymbol{S}$ is the second Piola-Kirchhoff stress tensor, $Q$ contains additional heat sources.

The simplified finite element model has significantly accelerated the simulation. However, the process is still relatively time-consuming and struggling to meet current expectations of battery pack design. Thus, a machine learning model coupled with physics information is needed further to accelerate the temperature prediction of the battery pack.

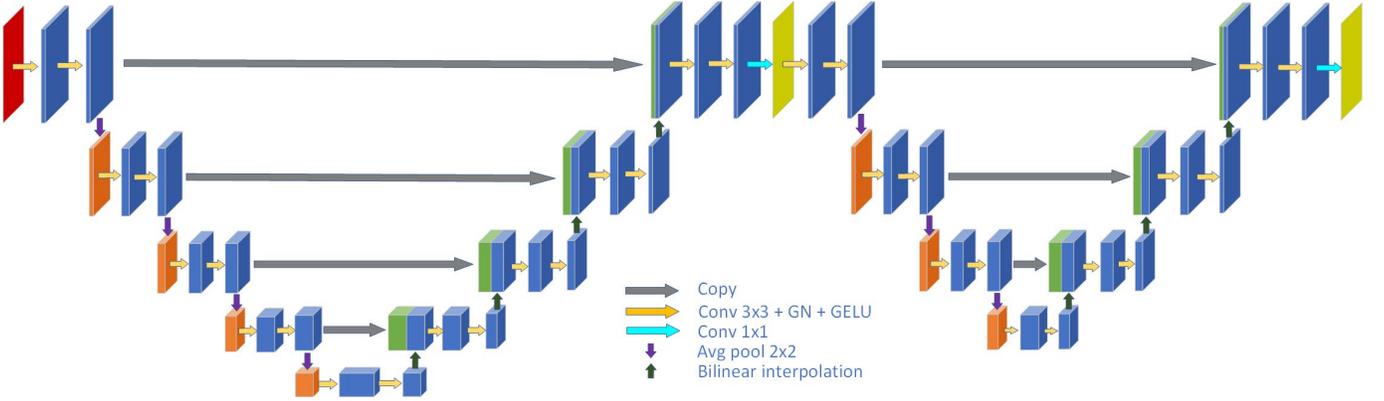

*Figure 2. Pipeline of multi-fidelity Physics-informed CNN.*

### 2.3 Finite Element Model Implementation

The physics information needs to be further simplified to integrate domain-specific knowledge from physics into the machine learning models. The steady state of the 2D temperature field can be simplified to Poisson's equation [12]:

$$\frac{\partial}{\partial x}\left(\lambda \frac{\partial T}{\partial x}\right) + \frac{\partial}{\partial y}\left(\lambda \frac{\partial T}{\partial y}\right) + \phi = 0, \quad (x,y) \in \Omega \quad (2)$$

where $\lambda(x,y)$, $\phi(x,y)$, and $T(x,y)$ stand for the thermal conductivity, intensity, and temperature distribution; $\Omega = \{(x,y) | 0 \leq x \leq a, 0 \leq y \leq b\}$ denotes the entire computation domain of battery pack; $a$ and $b$ are the length and width of battery pack, respectively.

For the proposed problem, the intensity of battery cells can be regarded as positive constants $\phi_b$, whereas the intensity of coolant is a negative value linearly related to the temperature. If $\Gamma \subset \Omega$ represents the domain of battery cells, the intensity distribution function can be written as

$$\phi(x,y) = \begin{cases} \phi_b, & (x,y) \in \Gamma \\ -k[T(x,y) - T_0], & (x,y) \notin \Gamma \end{cases} \quad (3)$$

where $k$ stands for the linear coefficient of coolant intensity; $T_0$ denotes the temperature of cold plates. In a battery pack, the thermal conductivity of battery cells and coolant displays different constants

$$\lambda(x,y) = \begin{cases} \lambda_b, & (x,y) \in \Gamma \\ \lambda_c, & (x,y) \notin \Gamma \end{cases} \quad (4)$$

where $\lambda_b$ and $\lambda_c$ represent thermal conductivity of battery cells and coolant, respectively. Given that all the side faces of battery pack are adiabatic, the boundaries of the simplified 2D problem should follow Neumann boundary conditions

The distribution function can be written as

$$\lambda \frac{\partial T(x,y)}{\partial n} = 0, \quad (x,y) \in \partial\Omega \quad (5)$$

where $\partial\Omega = \{(x,y) | x = 0, a \text{ or } y = 0, b\}$ denotes the boundary of battery pack; $n$ stands for the normal direction of boundary.

## 3 MULTI-FIDELITY PHYSICS-INFORMED CONVOLUTIONAL NEURAL NETWORK

### 3.1 Deep Multi-Fidelity Modeling

Since the intensity distribution $\phi(x,y)$, thermal conductivity distribution $\lambda(x,y)$, and the temperature field $T(x,y)$ can be meshed as 2D images, the steady-state temperature distribution estimation can be approached as an image-to-image regression problem within finite dimensional space, which can be addressed by CNNs. With the guidance of physics knowledge, CNNs can solve an entire family of heat transfer problems and realize unsupervised learning without temperature field observations. However, due to the exclusion of certain factors, such unsupervised learning with low-fidelity physics knowledge may deviate from the ground truth of the corresponding physical phenomenon. In addition, constrained by the computation efficiency of finite element simulation, high-fidelity data is markedly scarce compared with the amount of layout cases. To bridge the disparity between data and knowledge of various fidelities, a multi-fidelity semi-supervised physics-informed CNN with two-stage training is proposed as a surrogate model for battery temperature distribution estimation based on layout information.

As shown in Fig. 2, the proposed deep multi-fidelity model comprises two CNN modules: physics-informed backbone $\mathcal{B}(\cdot)$ and high-fidelity projection head $\mathcal{H}(\cdot)$. Given a full set of layout dataset $D^L = \{(\lambda_i, \phi_i)\}_{i=1}^M$, only a small subset contains the corresponding high-fidelity temperature field, denoted as $D^T = \{(\lambda_i, \phi_i), T_i\}_{i=1}^N$, where $N \ll M$. Thanks to the unsupervised training capability of physics-informed machine learning approach, the entire layout dataset $D^L$ without temperature information can be utilized to train the physics-informed backbone and learn the steady-state solutions of parametric PDEs in Eq. (2). The physics-informed loss function $\mathcal{L}_{phy}$ will be detailed in 3.2. At the end of the pre-training phase, the backbone can predict low-fidelity temperature based on layout data, i.e.

$$\hat{T} = \mathcal{B}(\lambda, \phi) \quad (6)$$

During the subsequent post-training stage, the high-fidelity projection head is appended to the trained physics-informed backbone, which serves as a base of prior knowledge. It should be noted that the parameter of backbone is fixed during this phase. With training on the temperature-labeled dataset $D^T$, the projection head refines the model output, compensating the gap between low- and high-fidelity estimations and producing a temperature distribution of enhanced fidelity. Upon completion of the post-training phase, the integrated multi-fidelity CNN effectively functions as the surrogate model for accurate battery temperature distribution estimation:

$$\tilde{T} = H(\mathcal{B}(\lambda, \phi)) \quad (7)$$

### 3.2 Physics-informed Backbone and Pre-training

Given a battery layout distribution, the physics-informed backbone CNN is designed to predict the 2D low-fidelity temperature field solution, $\hat{T}(x,y)$, as defined in Eq. (2), subject to the boundary conditions in Eq. (5). According to Eq. (8) and Eq. (4), both intensity and thermal conductivity distributions exhibit the same battery layout. However, as an implicit function of temperature, the coolant intensity remains undefined at the onset of training. Consequently, as shown in Fig. 2, only the thermal conductivity distribution $\lambda(x,y)$ is employed as the input of the CNN backbone, whereas the complete intensity distribution $\phi(x,y)$ is subsequently derived from the output of backbone and thermal conductivity distribution:

$$\phi(x,y) = \begin{cases} \phi_b, & \lambda(x,y) = \lambda_b \\ -k[\hat{T}(x,y) - T_0], & \lambda(x,y) = \lambda_c \end{cases} \quad (8)$$

After predicting the temperature field and full intensity distribution, a physics-informed loss function can be defined based on Eq. (2). First, by expanding the partial derivatives, Eq. (2) can be rewritten as

$$\frac{\partial \lambda}{\partial x}\frac{\partial T}{\partial x} + \frac{\partial \lambda}{\partial y}\frac{\partial T}{\partial y} + \lambda\left(\frac{\partial^2 T}{\partial x^2} + \frac{\partial^2 T}{\partial y^2}\right) + \phi = 0 \quad (9)$$

Considering a rectangular battery pack, the entire computation domain $\Omega$ can be discretized by m×n rectangular meshing. Assuming that the meshing steps h along the x-axis and y-axis are equal, i.e. $h = a/n = b/m$, Eq. (9) can then be approximated using finite difference method, i.e.

$$\frac{\lambda(x_{i+1},y_j)-\lambda(x_{i-1},y_j)}{2h} \cdot \frac{T(x_{i+1},y_j)-T(x_{i-1},y_j)}{2h} + \frac{\lambda(x_i,y_{j+1})-\lambda(x_i,y_{j-1})}{2h} \cdot \frac{T(x_i,y_{j+1})-T(x_i,y_{j-1})}{2h} + \frac{\lambda(x_i,y_j)}{h^2}[T(x_{i+1},y_j) + T(x_{i-1},y_j) + T(x_i,y_{j+1}) + T(x_i,y_{j-1}) - 4T(x_i,y_j)] + \phi(x_i,y_j) = 0 \quad (10)$$

where $(x_i, y_j)$ stands for the pixels on discretized computation domain. In this way, the physics-informed estimation error at each pixel can be written as

$$\delta_{phy}(x_i, y_j) = \left\|\hat{T}(x_i,y_i) - \frac{1}{4}\hat{T}'(x_i,y_i)\right\| \quad (11)$$

$$\hat{T}'(x_i,y_i) = \frac{h^2\phi(x_i,y_j)}{\lambda(x_i,y_j)} + \frac{\lambda(x_{i+1},y_j)-\lambda(x_{i-1},y_j)}{\lambda(x_i,y_j)} \cdot \frac{T(x_{i+1},y_j)-T(x_{i-1},y_j)}{4} + \frac{\lambda(x_i,y_{j+1})-\lambda(x_i,y_{j-1})}{\lambda(x_i,y_j)} \cdot \frac{T(x_i,y_{j+1})-T(x_i,y_{j-1})}{4} + T(x_{i+1},y_j) + T(x_{i-1},y_j) + T(x_i,y_{j+1}) + T(x_i,y_{j-1}) \quad (12)$$

where $\hat{T}(x_i, y_j)$ is the estimated low-fidelity temperature field, i.e. the output of physics-informed backbone; $\|\cdot\|$ stands for the $L_1$ norm.

To ensure all areas in the computation field converge at the same speed, a linear weighting factor is applied on the pixel level, formulated as

$$w_{phy}(x_i, y_j) = \eta_1 + \eta_2 \frac{\delta_{phy}(x_i,y_j)-\min(\delta_{phy})}{\max(\delta_{phy})-\min(\delta_{phy})} \quad (13)$$

where $\eta_1$ and $\eta_2$ are shifting and scaling factors; $\min(\delta_{phy})$ and $\max(\delta_{phy})$ are the minimum and maximum error among the entire domain. In this way, a weighted physics-informed loss function can be expressed as

$$\mathcal{L}_{phy} = \frac{1}{|\Omega|}\sum_{(x_i,y_j)\in\Omega} w_{phy}(x_i,y_j)\left\|\hat{T}(x_i,y_i) - \frac{1}{4}\hat{T}'(x_i,y_i)\right\| \quad (14)$$

Apart from the heat transfer equation, the boundary conditions should also be satisfied to determine the unique solution of the steady-state problem. For the Neumann boundary condition in Eq. (5), it can be discretized by central finite difference as

$$\begin{cases} \frac{\lambda}{2h}[\hat{T}(x_{n+1},y_j) - \hat{T}(x_{n-1},y_j)] = 0, & 0 \le j \le m \\ \frac{\lambda}{2h}[\hat{T}(x_{-1},y_j) - \hat{T}(x_1,y_j)] = 0, & 0 \le j \le m \\ \frac{\lambda}{2h}[\hat{T}(x_i,y_{m+1}) - \hat{T}(x_i,y_{m-1})] = 0, & 0 \le i \le n \\ \frac{\lambda}{2h}[\hat{T}(x_i,y_{-1}) - \hat{T}(x_i,y_1)] = 0, & 0 \le i \le n \end{cases} \quad (15)$$

Therefore, the reflect padding mode can be employed before computing physics-informed loss to enforce the padding values equal to the corresponding inner nodes.

The architecture of the physics-informed backbone employs a full-size UNet, which is adept at image-to-image regression problems. As an encoder-decoder framework, UNet captures global information by a contracting path whilst ensuring high-resolution predictions by a symmetric expanding path. As shown in Fig. 2, the input and output of UNet are discretized thermal conductivity $\phi(x,y)$ and low-fidelity temperature estimation $\hat{T}(x,y)$, respectively. The encoder comprises five convolution blocks. Each block contains two groups of 3×3 convolution layers with group normalization (GN) and GELU activation function, which increases the non-linearity of the CNN backbone. A 2×2 average pooling filter with stride 2 is also appended to downsample the size of feature maps. The size and channels of feature maps are labeled at the bottom and top of each block in Fig. 2.

Symmetrically, the decoder comprises five convolution blocks starting with a bilinear interpolation layer for upsampling and a concatenation of feature maps in the same level. Then, the same 3×3 convolution blocks with GN and GELU activation function are utilized for feature extraction. At the end of decoder, a 1×1 convolution layer is employed to output the low-fidelity feature map. To ensure the boundary conditions in feature maps, the reflect padding mode is applied for all convolution operators.

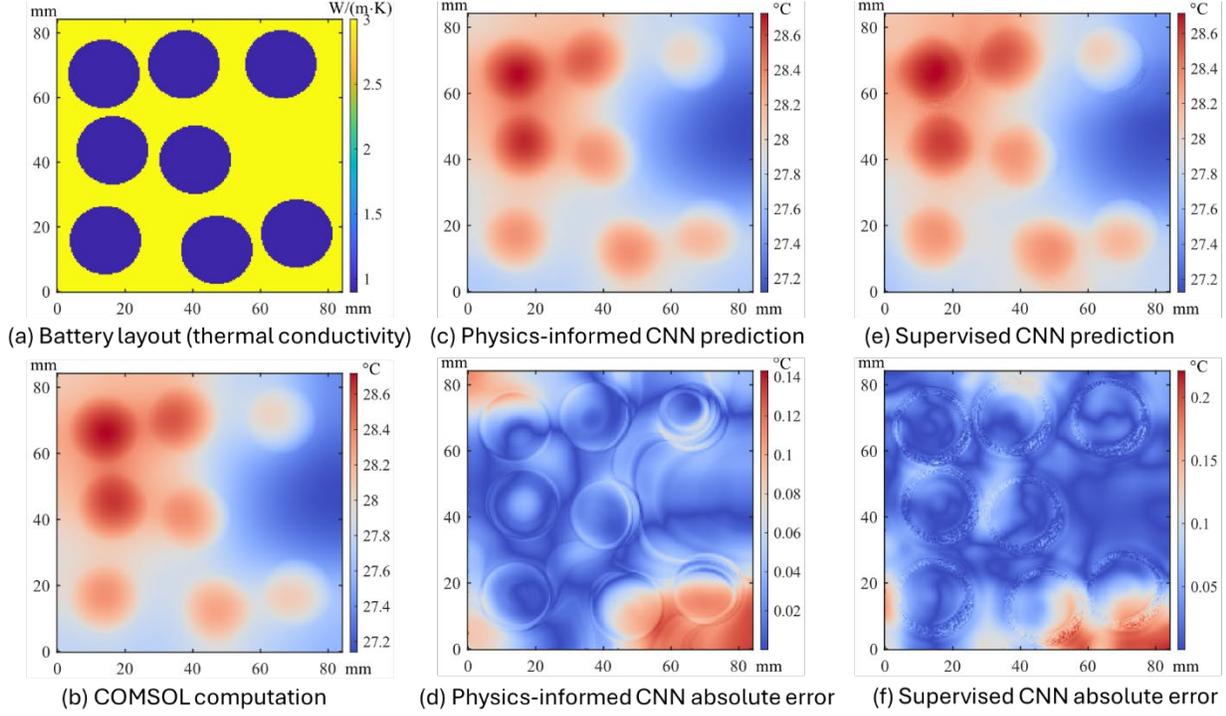

*Figure 3. Battery temperature distribution prediction by multi-fidelity physics-informed CNN and supervised CNN.*

### 3.3 High-fidelity Projection Head and Post-training

The high-fidelity projection head is appended to the end of physics-informed backbone after the pre-training is completed. Given that the backbone has already incorporated substantial physics information, the non-linearity between high- and low-fidelity temperature predictions is considerably reduced compared to that of the backbone. Therefore, this paper introduces a reduced-size UNet as the projection. As shown in Fig. 2, the reduced UNet comprises only four levels of convolution blocks, each consisting of a 3×3 convolution layer with reflect padding, group normalization, and ReLU activation function. It should be noted that the ReLU activation is utilized, instead of GELU, to reduce the non-linearity of the projection head. In addition, the average pooling and bilinear interpolation are applied to shrink or expand the size of feature maps between different levels.

The post-training of high-fidelity projection head is a data-driven phase, during which the physics-informed backbone is fixed, and only the temperature-labeled dataset $D^T$ is considered for loss function computation. In order to balance the convergence speed among computation areas, the linear weighting strategy is also applied in the supervised post-training phase. Given the high-fidelity temperature field estimation $\tilde{T}$ and its ground truth $T$, the high-fidelity estimation error at each pixel is defined as

$$\delta_{data}(x_i, y_j) = \|\tilde{T}(x_i, y_j) - T(x_i, y_j)\| \quad (16)$$

Then, the weighted data-driven loss function can be computed as

$$\mathcal{L}_{data} = \frac{1}{|\Omega|}\sum_{(x_i,y_j)\in\Omega} w_{data}(x_i, y_j)\|\tilde{T}(x_i, y_j) - T(x_i, y_j)\| \quad (17)$$

with

$$w_{data}(x_i, y_j) = \eta_1 + \eta_2 \frac{\delta_{data}(x_i,y_j) - \min(\delta_{data})}{\max(\delta_{data}) - \min(\delta_{data})} \quad (18)$$

After post-training, deep the multi-fidelity model can be obtained for accurate battery temperature distribution estimation.

## 4 EXPERIMENT VALIDATION

### 4.1 Dataset and Experimental Settings

In our experiment, 8 battery cells are placed in a box with square cross-section. The diameter of each battery cell is 21 mm and the size of heat transfer area is 84 mm × 84 mm. A gap of more than 2 mm is maintained between different batteries and between batteries and boundaries. 2000 cases with different battery layout are generated, which is split into 1000, 500, and 500 for training, validation, and testing. Each data pairs contain a thermal conductivity map, an initial intensity map, and a corresponding steady-state temperature distribution obtained by COMSOL simulation as ground truth. The layout and temperature distribution are uniformly meshed into 200 × 200 grids. Assuming that the height of battery cells is 70 mm in the

3D problem, the intensity of battery cells and linear intensity parameter of coolant are $\phi_b = 12348.35$ W/m$^2$ and $k = 3000$ W/(m$^2$·K), which align with the settings in 2.2. The thermal conductivity of battery cells and coolants are $\lambda_b = 0.89724$ W/(m·K) and $\lambda_c = 3$ W/(m·K), respectively. The temperature of cold plates is $T_0 = 25$°C. It should be noted that the coolant intensity in initial intensity maps is set to zero since it is implicitly related to the temperature.

For the proposed multi-fidelity physics-informed CNN, all the 1000 layout maps in the training dataset are utilized in pre-training. During the post-training phase, only 100 cases with simulated temperature distributions in the training dataset are utilized for supervised learning. The hyperparameters of weighted loss function are set to $\eta_1 = 0$ and $\eta_2 = 10$. The Adam optimizer with an initial learning rate of 0.001 is employed during the pre- and post-training phases. The learning rate decays exponentially with a shrinking factor of 0.85. The batch size is set to 1, and the epoch during the pre-training and post-training phases are set to 10 and 15, respectively. To demonstrate the efficiency of the proposed method in small-sample learning, a full-size UNet is trained, as a comparison, based on the supervised learning scheme using the same amount of labeled data. All models are performed on a single NVIDIA Tesla P100 GPU.

### 4.2 Results and Discussions

Fig. 3 demonstrates a test case for battery temperature distribution prediction using the proposed multi-fidelity physics-informed CNN and its comparison with supervised CNN. All temperature distributions are presented in degrees Celsius (°C). As depicted in Fig. 3(c), the proposed model accurately predicts the temperature distribution in the battery pack, closely aligning with the COMSOL computations. The maximum and minimum temperature predictions correspond well with those derived from finite element simulations. Fig. 3(d) reveals that the estimation error of the proposed method remains below 0.14 °C, with over 80% of the computation areas exhibiting errors under 0.08 °C, manifesting great power by fusing physics knowledge and data. In contrast, while the supervised UNet shows similar temperature distribution prediction, it yields a maximum absolute error exceeding 0.2 °C, approximately 4.5 times higher than that of the proposed method. Additionally, without physics knowledge, the supervised CNN generates rougher temperature distribution estimation, especially around the battery-coolant interface, when lacking labeled data.

In order to quantify the prediction performance, four indices - the mean absolute error among while domain (MAE), the mean absolute error on battery cells (BMAE), the maximum absolute error among while domain (Max-AE), and the absolute error of the maximum temperature (MT-AE), are introduced for model comparison, as listed in Table 1. The metrics demonstrate that the physics-informed CNN consistently outperforms the supervised CNN. Specifically, the MAE and BMAE of PI-CNN are 17% and 18% lower, respectively, than those of supervised CNN, indicating superior overall prediction accuracy. The Max-AE and MT-AE of PI-CNN are reduced by 15% and 2.3%, respectively, manifesting the enhanced capability in addressing extreme cases. These results underscore the promising potential of the proposed multi-fidelity physics-informed CNN for optimal design and thermal management control of battery packs.

Table 1 – Quantitative comparison of prediction performance

| Method | MAE | BMAE | Max-AE | MT-AE |
|---|---|---|---|---|
| PI-CNN | 0.0360 | 0.0345 | 0.2045 | 0.0463 |
| Supervised CNN | 0.0434 | 0.0422 | 0.2403 | 0.0474 |

### 5 CONCLUSION

This paper proposes a multi-fidelity physics-informed CNN for battery temperature distribution prediction. First, a 3D battery cooling problem is simplified to a 2D heat transfer scenario, which is simulated using finite element tools. Then, domain-specific knowledge is incorporated into deep learning models through a physics-informed loss function that follows the low-fidelity heat transfer PDEs. The UNet CNN with a multi-fidelity projection head is introduced as surrogate model, predicting the battery temperature distribution with a small amount of labeled dataset. Results indicate that the proposed method enhances temperature distribution estimation accuracy by 15% over traditional data-driven approaches with comparable limited training datasets. Future research will concentrate on battery optimal design and control leveraging physics-informed machine learning techniques.

### ACKNOWLEDGEMENTS

This work was partially supported by the National Science Foundation through Engineering Research Center for Power Optimization of Electro-Thermal Systems (POETS) under Cooperative Agreement No. EEC-1449548 and through the award CMMI-2037898.

*BIOGRAPHIES*

Zheng Liu, PhD Candidate
Department of Industrial and Enterprise Systems Engineering
University of Illinois Urbana-Champaign
104 S Mathews Ave
Urbana, Illinois 61801 USA

e-mail: zhengl6@illinois.edu

Zheng Liu is a Ph.D. candidate at University of Illinois Urbana-Champaign, Urbana, IL, USA. He received his M.S. in Mechanical Engineering from Cornell University, Ithaca, NY, USA. He is an IEEE, ASME, and IISE student member. His research interests focus on generative design, physics-informed machine learning, reliability-based design optimization, control co-design optimization, and life cycle assessment, with an emphasis on applications including battery thermal management systems, power electronics, and additive manufacturing.

Yuan Jiang, PhD Student
Department of Industrial and Enterprise Systems Engineering
University of Illinois Urbana-Champaign
104 S Mathews Ave
Urbana, Illinois 61801 USA

e-mail: yuanj5@illinois.edu

Yuan Jiang received the B.S. and M.S. degrees in vehicle engineering from Tongji University, Shanghai, China, in 2020 and 2023, respectively. He is currently working toward the Ph.D. degree in industrial engineering with the University of Illinois Urbana-Champaign, Urbana, IL, USA. His research interests include prognostics and health management, fault diagnosis and condition monitoring, as well as reliability-based design optimization, using physics-informed machine learning and digital twin techniques.

Yumeng Li, PhD
Department of Industrial and Enterprise Systems Engineering
University of Illinois Urbana-Champaign
104 S Mathews Ave
Urbana, Illinois 61801 USA

e-mail: yumengl@illinois.edu

Yumeng Li is currently an Assistant Professor in the Department of Industrial and Enterprise Systems Engineering, University of Illinois Urbana-Champaign, Urbana, IL, USA. She earned her doctoral degree in the Department of Aerospace Engineering at Virginia Tech in 2014. Professor Li's research interests include uncertainty quantification in multiscale simulation framework, physics-based machine learning, Surrogate modeling, multiscale Simulation, and integrated computational material engineering.

Pingfeng Wang, PhD
Department of Industrial and Enterprise Systems Engineering
University of Illinois Urbana-Champaign
104 S Mathews Ave
Urbana, Illinois 61801 USA

e-mail: pingfeng@illinois.edu

Pingfeng Wang is currently a Professor and the Jerry S. Dobrovolny Faculty Scholar with the Department of Industrial and Enterprise Systems Engineering, University of Illinois Urbana-Champaign, Urbana, IL, USA, and an affiliate Professor with Materials Research Laboratory, Urbana, IL, USA. His research interests include engineering system design for reliability, failure resilience and sustainability, and prognostics and health management. Dr. Wang is a member of IEEE, ASME, and AIAA. He was the President of the Quality Control and Reliability Engineering (QCRE) division at IISE, the executive committee for ASME design automation society, and on the Board of Directors for Reliability and Maintainability Symposium (RAMS).